\documentclass[ijoc,sglanonrev]{informs4}
\usepackage{eqndefns-left} 
\RequirePackage{tgtermes}
\RequirePackage{newtxtext}
\RequirePackage{newtxmath}
\RequirePackage{bm}
\RequirePackage{endnotes}

\OneAndAHalfSpacedXII 

\usepackage{algorithm}
\usepackage{algpseudocode}
\usepackage{tikz}
\usepackage{ulem}

\usepackage{caption}
\usepackage[labelfont=sf]{subcaption}
\usepackage{natbib}
 \bibpunct[, ]{(}{)}{,}{a}{}{,}%
 \def\bibfont{\small}%
 \def\BIBand{and}%

\EquationsNumberedThrough    

\TheoremsNumberedThrough     
\ECRepeatTheorems  %

\MANUSCRIPTNO{IJOC-0001-2024.00}

\usepackage{geometry}
\begin{document}


\RUNAUTHOR{Benfield et al.}

\RUNTITLE{Classification under strategic adversary manipulation using pessimistic bilevel optimisation}


\TITLE{Classification under strategic adversary manipulation using pessimistic bilevel optimisation}

\ARTICLEAUTHORS{%
\AUTHOR{David Benfield\textsuperscript{\textit{a\footnote{The work of this author was supported by an EPSRC Studentship with reference 2612869.}}}, Stefano Coniglio\textsuperscript{\textit{b}},
Martin Kunc\textsuperscript{\textit{c}},
Phan Tu Vuong\textsuperscript{\textit{a}},
Alain Zemkoho\textsuperscript{\textit{a\footnote{The work of this author was supported by an EPSRC grant with reference EP/X040909/1.}}}}
\AFF{\textsuperscript{\textit{a}}School of Mathematical Sciences, University of Southampton, \EMAIL{\{db3g17,\, a.b.zemkoho,\, t.v.phan\}@soton.ac.uk}; \textsuperscript{\textit{b}}Department of Economics, University of Bergamo, \EMAIL{stefano.coniglio@unibg.it};
\textsuperscript{\textit{c}}Southampton Business School, University of Southampton, \EMAIL{m.h.kunc@soton.ac.uk}}}



\ABSTRACT{%
Adversarial machine learning concerns situations in which learners face attacks from active adversaries.
Such scenarios arise in applications such as spam email filtering, malware detection and fake-image generation, where security methods must be actively updated to keep up with the ever improving generation of malicious data. We model these interactions between the learner and the adversary as a game and formulate the problem as a pessimistic bilevel optimisation problem with the learner taking the role of the leader. The adversary, modelled as a stochastic data generator, takes the role of the follower, generating data in response to the classifier.
While existing models rely on the assumption that the adversary will choose the least costly solution leading to a convex lower-level problem with a unique solution, we present a novel model and solution method which do not make such assumptions. We compare these to the existing approach and see significant improvements in performance suggesting that relaxing these assumptions leads to a more realistic model.
}%





\maketitle


\section{Introduction}\label{sec:Intro}

Adversarial learning is the subfield of machine learning that involves the
understanding and countering of the adaptive strategies an adversary may put in place in response to the deployment of a security measure \citep{adv_book_2023}. By studying how adversaries evolve their behaviour, this discipline allows for the design of robust and resilient systems to mitigate evolving threats \citep{adv_deep_learning_survey, GameTheorySurvey}, 
 for example designing protection from credit card fraud \citep{INFORMS_credit_card_fraud} or finding the optimal placement of air defence systems \citep{INFORMS_defense}.
%
In many cases, security measures need to be constantly updated to keep up with the ever evolving adversary. One such example is the case of spam email filtering, where spammers might simply change how they write their emails to evade detection by
filters \citep{Adv_Class}. When data are changed in response to the deployment of a new classifier, the distribution of the data will be different from that of the data the classifier was trained on, resulting in new vulnerabilities and posing security threats to its users. The United States Federal Bureau of Investigation has estimated that close to \$12 billion was lost to email scams between 2013 and 2018 \citep{FBI}. In the literature surrounding adversarial machine learning, these attacks are commonly referred to as \textit{exploratory} attacks \citep{taxonomy_original}. Similar attacks might occur in security scenarios such as malware detection \citep{Malware_Detection} and network intrusion traffic \citep{Network_Intrusion}. In a similar vein, and more recently, vulnerabilities in deep neural networks (DNN) are being discovered, particularly in the field of computer vision and image classification; small perturbations in the data can lead to incorrect classifications by the DNN \citep{DNNAttacks1, DNNAttacks2}. These vulnerabilities raise concerns about the robustness of the machine learning technology that is being adopted and, in some cases, in how safe relying on their predictions could be in high-risk scenarios such as autonomous driving \citep{DNNSelfDriving} and medical diagnosis \citep{DNNMedical}.

In this work, we focus on the use of game theory to model these attacks in order to train resilient classifiers. Typically, these approaches construct a competition between two players: the learner, whose goal is to train a classifier, and the adversary, who manipulates data with the goal of disrupting the classifier, see for example \citep{Adv_Class} for early work in modelling adversarial learning. The precise structure of such a model then depends on a number of factors, such as whether the attack occurs either at training time \citep{Bruck_Ext}, or implementation time \citep{Brückner_Scheffer_2011}. Further to this, while some games assume the players act simultaneously \citep{NE_static, Deep_learning_Games}, others allow for sequential play where either the adversary acts first \citep{Liuetall, AdvLeadExt, AdvLeadExtTwo, kantarcıoğlu_xi_clifton_2010} or the learner acts first \citep{taxonomy, Brückner_Scheffer_2011}. Further information on the use of game theory in adversarial learning can be found in Section \ref{sctn:litrev} below.

Our attention in this article is focused on attacks at implementation time and since the attacks are made after the classifier has been established, we model the game as a bilevel optimisation problem with the learner as the leader. Under this formulation, the learner acts first to train a classifier, after which, the adversary generates new data while observing the classifier's capabilities. Within the bilevel optimisation model, we make the pessimistic assumption, meaning that in the case of multiple optimal solutions, we assume the worst case for the learner, the best strategy for the learner is then to train the classifier that minimises the amount of damage the adversary may cause.


A pessimistic bilevel formulation with the learner as leader has previously been investigated by \cite{Brückner_Scheffer_2011} and we extend their work in several ways. Similar to their model, we work under the assumption that the adversary's data is continuous. In their work, this was ensured by making the assumption that the data are passed through some feature map, such as principal component analysis (PCA). Such an assumption ensures categorical variables, such as binary bag-of-words represents of text, can be represented in a continuous space. Further to this, the authors of \cite{Brückner_Scheffer_2011} argue that in the event of multiple optimal solutions, all optimal solutions will be represented by the same point in the map-induced feature space, an element that played a key role in their solution method. In this work, we keep the data in its original space, allowing for distinct multiple optimal solutions. We achieve this by modelling the adversary as a stochastic player who samples data from some parameterised distribution. Under the motivation of representing text-based data through binary bag-of-words vectors, we construct a continuous approximation of the distribution for binary data.

To measure the success of the adversary's data, the authors of the previous paper propose a loss function combined with a quadratic regularisation term which penalises on the size of the transformation on the data, measured by the distance from its original position. They argue that the adversary has an incentive to choose the transformation that requires the least work. Under the additional assumption that the loss function is convex, the resulting objective function must be strictly convex due to the quadratic regularisation. Assuming that the map-induced feature space is unrestricted, this guarantees a unique solution in the map-induced feature space, another key element to their solution method. In this work, we make no assumption on the convexity of the objective functions or the uniqueness of the solution of the lower-level problem.

With a unique solution in the map-induced space, and all optimal solutions in the original space sharing the same representation in the map-induced space, the pessimistic perspective is removed from the bilevel optimisation problem. The assumption of lower-level convexity further allowed the authors to solve the bilevel optimisation problem by finding a point which satisfies its corresponding Karush-Kuhn-Tucker (KKT) conditions. Since we make no assumptions about the convexity of the lower-level problem, we instead make use of the value function reformulation which relies on the definition of the lower-level problem, and does not require such an assumption. From the value-function reformulation, we build a system of equations and use the Levenberg-Marquardt method to solve. We demonstrate through experiments that our model accurately captures adversarial influence on spam emails and fake reviews and that the solution method converges to a solution which out-performs the current pessimistic bilevel optimisation model by \cite{Brückner_Scheffer_2011}, providing more accurate predictions on future data.

The key contributions of this paper can be summarised as follows. We remove the assumption that the data have been passed through some feature map and instead construct an adversary that samples data from some distribution in order to ensure continuous variables while also remaining in the original feature space. Further to this, we make no assumption on the convexity of the lower-level problem or the uniqueness of its solution, thus preserving distinct multiple optimal solutions. Then, to solve the pessimistic bilevel optimisation model in absence of these assumptions, we propose a tractable and efficient method that assumes only that the objective functions are twice continuously differentiable. Finally, we demonstrate in numerical experiments that our model accurately captures the adversarial nature of text--based data and that the solutions found outperform the current pessimistic approach to adversarial classification.

The paper is structured as follows. Relevant contributions from the literature are summarised in Section \ref{sctn:litrev}. In Section \ref{sctn:Protected Training}, we construct a novel pessimistic bilevel program to model implementation-time attacks to binary classifiers and a corresponding solution method in Section \ref{sctn:Solution Method}. We present the results from some numerical experiments in Section \ref{sctn:Applications} before 
concluding the paper in Section \ref{sctn:Conclusion}.

\section{Previous and related works}\label{sctn:litrev}

The literature surrounding adversarial machine learning is vast, and in this section we focus specifically on reviewing the relevant
use cases of game theory to model adversarial classification scenarios. For an in-depth study which is not limited just to game theory, the reader is referred to the comprehensive book by \cite{adv_book_2023}.

Classification games in adversarial learning typically define one player to be the learner, whose goal is to train a classifier and the other as an adversary, who manipulates data with the goal of disrupting the classifier.
The players are typically defined by their set of strategies, such as the weights of the classifier for the learner or a set of data transformations for the adversary, and corresponding utilities, such as loss functions, which measure the success of these strategies. See for example \cite{Adv_Class} for early work in modelling adversarial learning.

The nature of the attack depends on the stage of the classifier's development at which it takes place and \cite{taxonomy_original} developed a taxonomy to help categorise attacks in such a way.
In particular, they separate attacks that are made at training time to those made at test or implementation time. For training-time attacks, we might model an adversary who influences the output of a classifier by strategically modifying its training data; see, for example, \cite{Bruck_Ext}. In this work, however, we focus on the case of attacks at implementation time where the adversary seeks to evade the classifier. This is suitable, for instance, for the case of a spammer modifying their emails to evade detection by an email filter.

We can further categorise the literature by whether the player's strategies are decided simultaneously, see e.g. \citep{NE_static, Deep_learning_Games} or sequentially. 
Most approaches argue that a sequential game provides a more realistic representation since it allows one player to observe the other's move before making their own. In solving these games, we can identify strategies for the leading player while anticipating how the following player will react. These sequential games can then be categorised based on which player takes on the role of the leader. For games with the adversary as leader, see,  for example, \cite{Liuetall, AdvLeadExt, AdvLeadExtTwo, kantarcıoğlu_xi_clifton_2010}. In these games, since the adversary makes the first move, they would not be able to observe the learner's strategy when deciding on their best course of action, and would instead have to anticipate how the learner will react.
In our work we set the learner as the leader as we model scenarios where the learner is likely to publish their classifier publicly. For example, the email filter will have already been established when the spammer starts modifying their emails.

To further divide the literature, we separate zero-sum games, where player's utilities sum to zero, from non-zero-sum games where they do not. Under a zero-sum game, where the gain of one player comes directly at the cost of the other, Nash equilibria can be found by solving for the minimax outcome. For zero-sum approaches, see, for example, \cite{ACRE, Adv_Class, ACRE_ext, Nightmare_at_test, Startegic_class}.
However, \cite{Brückner_Scheffer_2011} argue that such models are often overly pessimistic since the adversary does not necessarily have motives that are purely antagonistic in nature. They give the example of a fraudster, whose goal of maximising profit from phished account information is not necessarily the inverse of an email filter's goal of maximising the rate of correctly filtered spam emails. A non-zero-sum game with separately defined objectives could provide a more realistic representation.


Due to its hierarchical structure, a non-zero-sum sequential game can be modelled as a bilevel optimisation problem, where the solution to the lower-level (follower's) problem is dependent on the solution to the upper-level (leader's) problem. Since the follower can observe the leader's move before making their own, the leader must anticipate how the follower will react.
In the case of multiple optimal strategies for the follower, we must make some assumption about how cooperatively they will behave and so which one they choose. In the literature on bilevel optimisation, this decision coincides with making either the optimistic or the pessimistic assumption \citep{Bilevel_Optimisation_Adv_and_Chal}. In the optimistic setting, we assume the lower-level player has an incentive to select strategies that benefit the upper-level player, for example, \cite{Bruck_Ext} utilise such an assumption to model training-time attacks. Under this model, where the learner takes on the role of the follower, it might be reasonable to assume no antagonistic intent. However, for attacks at implementation time, where the adversary takes on the role of the follower, this is no longer the case, as suggested by \cite{Brückner_Scheffer_2011}. We recall that, in the case where the lower-level problem possesses a unique solution, we get a special cooperative model. Formulations of such a nature are constructed or studied in \cite{Liuetall} and \cite{Wangetal}, for example.

A pessimistic model with the learner as leader was proposed by \cite{Brückner_Scheffer_2011} which in recent years has been extended, such as by \cite{Bruck_Ext} who consider attacks to the training set and \cite{bruck_Ext_2} who construct randomised prediction games.

To solve constrained pessimistic bilevel optimisation problems, many methods transform the problem into a mathematical program with equilibrium constraints by first replacing the lower-level problem with its Karush-Kuhn-Tucker (KKT) conditions, see, e.g. \cite{Brückner_Scheffer_2011} who utilise such a method to solve the pessimistic model in an adversarial setting. However, these methods often rely on frameworks which require some strong assumptions such as a convex lower-level problem. Whereas the alternative optimal-value reformulation relies only on the definition of the lower-level problem.
For further information regarding solution methods to pessimistic bilevel optimisation problems and bilevel optimisation problems more generally, the reader is referred to the comprehensive surveys conducted in \cite{Bilevel_Optimisation_Adv_and_Chal}.
For works on solving pessimistic bilevel optimisation problems arising in a game-theoretic settings with many lower-level followers, we refer the reader to \cite{coniglio2020computing,castiglioni2019leadership}.
The close relationship between bilevel optimisation, robust optimisation, and stochastic programming is explored in the survey~\cite{bolusani2020unified}, for example.

\section{Protected training against adversarial attacks}\label{sctn:Protected Training}
We model adversarial attacks at implementation time as a game between two players: the learner, whose objective is to train a classifier to detect adversarial data, and the adversary, whose objective is to create data that evades detection by the classifier. In this section, we develop the pessimistic bilevel optimisation problem to model this game. In a bilevel program, the lower-level's optimal solution is dependent on the solution found for  the upper-level problem. As such, we refer to the lower-level problem as the follower's problem and to the upper-level as the leader's problem. Since we focus on attacks at time of implementation, we assume that the classifier has already been trained when the attack takes place,
and more importantly, that the adversary might observe the capabilities
of the classifier when creating their data. For this reason, it seems sensible to place the adversary in the lower level.
From the learner's perspective, solving this game then equates to finding a classifier that anticipates changes in the distribution.

We make the assumption that the adversary will not act cooperatively, meaning that in the case of multiple optimal solutions to the adversary's (lower-level) problem, the learner assumes that the adversary will choose the worst one for the learner. The resulting problem is known in the literature as a pessimistic bilevel optimisation problem. In the adversarial learning setting, such a pessimistic problem with the learner as leader has previously been investigated by \cite{Brückner_Scheffer_2011}. In their work, the lower-level variable is passed through a feature map, such as principal component analysis, resulting in an unrestricted lower-level feasible region. A quadratic regularisation term is placed on the lower-level objective function to penalise on large modifications carried out by the adversary on their data.
The idea is that the adversary has an incentive to choose the transformation that requires the smallest amount of work. This regularisation term ensures a strictly convex objective function, and \cite{Brückner_Scheffer_2011} build on the fact that if the objective is defined over the unrestricted vector space, which is a convex set, the lower-level problem must posses a unique solution. In our work, we do not make any assumptions about the convexity of the lower-level objective function or the uniqueness of the solution to lower level-problem.

Let $X \in \mathbb{R}^{n \times p}$ be the static set of $n \in \mathbb{N}$ instances of data (one per row of $X$) containing $q \in \mathbb{N}$ features each, and let
$y \in \{0,1\}^{n}$ be the corresponding collection of binary classes. 
Suppose the adversary wishes to create
a set of data containing $m \in \mathbb{N}$ samples of $q$ features.
Representing text-based data, such as spam emails, often requires discrete features, such as binary bag-of-words where data are represented by binary vectors. Since having a bilevel optimisation model with continuous variables is more convenient to scale up numerical algorithms, rather than letting the adversary manipulate the data directly, we model the adversary as a stochastic player who samples their data from some distribution. For some pre-defined uniformly distributed $z \in (0,1)^{m \times q}$, where $z_{ij} \sim U(0,1)$ for all $i \in \{1,\dots,m\}$ and $j \in \{1,\dots q\}$, we develop a generator function $G : (0,1)^q \times \mathbb{R}^{2q} \rightarrow \mathbb{R}^{q}$ which works by taking as input a row $z_i$ for $i \in \{1,\dots,m\}$ of $z$ and maps these to adversarial data by sampling from the inverse cumulative distribution function (CDF) parameterised by $\theta$. For some random variable $V$, the CDF, $\Gamma$, is defined as $\Gamma(v;\theta) := P(V\leq v)$, for some $v \in \mathbb{R}$. By allowing the adversary control over $\theta$, they can find the optimal distribution to sample data that evades the learner's classifier.

Assuming the learner knows the adversary's data, their task is to use the combination of static and adversarial data
to train a classifier. With weights $w \in \mathbb{R}^q$, the learner makes classifications through their prediction function $\sigma: \mathbb{R}^{q} \times \mathbb{R}^{q} \rightarrow [0,1]$, which 
maps $X_i$ to the learner's estimate of $X_i$ belonging to class $1$. To find the optimal weights of $\sigma$, the learner
seeks to minimise some twice continuously differentiable loss function $\mathcal{L}: \mathbb{R}^{q} \times \mathbb{R}^{q} \times \{0,1\} \rightarrow \mathbb{R}$ which penalises on incorrect predictions. For a sample of data $X_i$ with corresponding class $y_i$, the loss measured for weights $w$ is given by $\mathcal{L}(w, X_i, y_i)$. We do not assume that the adversary necessarily only produces one class of data. Let $\gamma$ be the true class of the adversary's data, the learner's objective function is then defined to be the sum of the loss on the static data and the generated data: 
\begin{equation}
\label{eqn:UpperLevel}
    F(w,\theta) := \sum_{i=1}^{n} \mathcal{L}(w, X_i, y_i) + \sum_{i = 1}^m \mathcal{L}(w, G(z_i, \theta), \gamma_i).
\end{equation}
The adversary seeks to generate data that evades the learner's classifier. We define a separate twice continuously differentiable loss function $\ell: \mathbb{R}^{q} \times \mathbb{R}^{q} \times \{0,1\} \rightarrow \mathbb{R}$ for the adversary. For adversarial sample $G(z_i, \theta)$ with true class label $\gamma_i$, the value of $\ell(w, G(z_i, \theta), \gamma_i)$ measures how well the adversary's data was able to fool the classifier, such as by penalising when the learner's classifier correctly identifies the true calss of the adversary's data. We define the adversary's objective function to be the sum over the loss function on the adversary's data: 
\begin{equation}
    \label{eqn:lower}
    f(w, \theta) := \sum_{i = 1}^m \ell(w,G(z_i, \theta), \gamma_i).
\end{equation}

Since we model problems where attacks happen at implementation time, we let the learner act first. Then, given some weights $w$ found by the learner, the adversary seeks to minimise their loss function to find the optimal distribution parameters to create new data that can fool the learner. We define this set of optimal distribution parameters as
\begin{equation}\label{eqn:solution_map}
    S(w) := \argmin_{\theta \in \mathbb{R}^{2q}} f(w,\theta),
\end{equation}
which forms the lower-level optimisation problem of the bilevel optimisation model. It is possible that the lower-level problem possesses multiple optimal solutions, and so we need to make some assumption about which one the adversary chooses.
In the optimistic case, we would assume that the solution picked by the adversary would be the one which most benefits the learner, i.e., the solution that minimises $F$ with respect to $\theta$. However, we make the assumption that the adversary is more likely to choose the solution that does the maximum harm to the learner. Hence, we take a pessimistic approach to this problem and assume the adversary chooses the solution that maximises the learner's objective function. The complete bilevel optimisation problem
is then defined as
\begin{equation}\label{eqn:Upper Level}
    \min_{w \in \mathbb{R}^{q}} \max_{\theta \in S(w)} F(w,\theta).
\end{equation}

This formulation, with objective functions defined as in \eqref{eqn:UpperLevel} and \eqref{eqn:lower}, provides a general, non-zero-sum, sequential approach to adversarial attacks at implementation time that makes no assumptions on the convexity of the lower-level problem or the uniqueness of its solution. In solving this bilevel optimisation model, the learner will anticipate how the adversary will react to their classifier, and so find one that minimises the amount of damage the adversary can do.

To sample adversarial data, we can define $G$ to be the inverse CDF of some parameterised distribution. Since the inverse CDF of commonly used distributions do not posses a closed-form continuous expression, we work to develop an approximation for binary variables under the motivation that text-based data, such as spam emails, will be represented by binary bag-of-words vectors.

\begin{figure}[]
\FIGURE
{
\subcaptionbox{Varying $\alpha$ ($\beta = 0.5$)\label{fig:VaryingAlpha}}
{\includegraphics[width=0.5\textwidth]{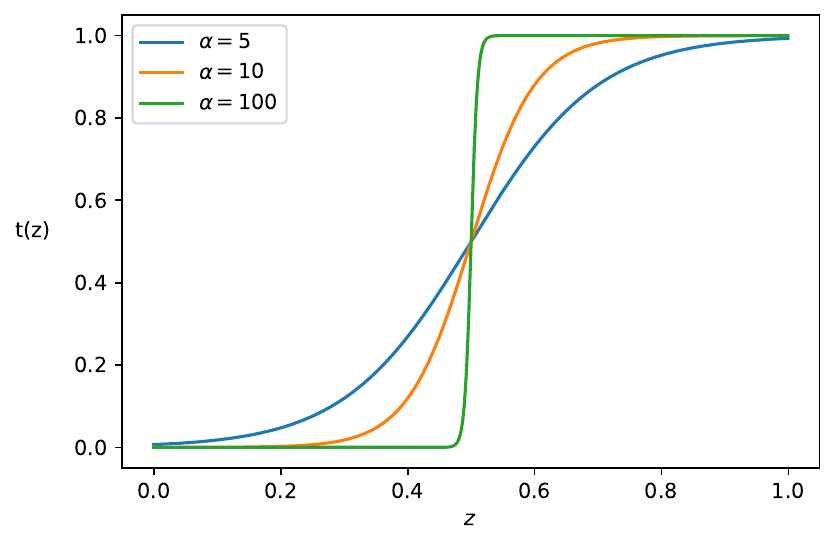}}
\subcaptionbox{Varying $\beta$ $(\alpha = 100)$\label{fig:VaryingBeta}}
{\includegraphics[width=0.5\textwidth]{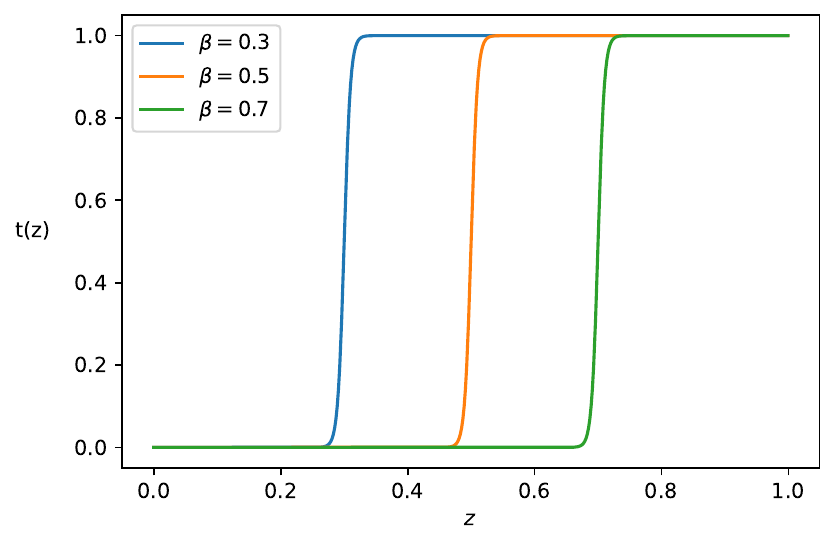}}
}
{
{\textmd{Varying parameters of the tanh approximation}}\\
\label{fig:VaryingParameters}}
{
}
\end{figure}

Let $\Gamma : \mathbb{R} \times \mathbb{R} \rightarrow [0,1]$ be the parameterised marginal CDF of some binary feature.
Then its inverse, $\Gamma^{-1} : [0,1] \times \mathbb{R} \rightarrow \mathbb{R}$ can be expressed by a step function which breaks at the proportion of each value of the feature. For a binary feature with values $0$ and $1$, let $\beta \in [0,1]$ be the proportion of value $0$; then, the inverse CDF of this feature is defined as
\begin{equation}
\label{eqn:StepFunction}
    \Gamma^{-1}(v, \beta) := \begin{cases} 
      0 & \mbox{ if }\, v\leq \beta, \\
      1 & \mbox{ otherwise},
   \end{cases}
\end{equation}
for $v \in [0,1]$.
To sample a new instance of this binary feature, we can first sample $v$ from the uniform distribution, $v \sim U(0,1)$, and map it to an instance of the binary feature by $\Gamma^{-1}(v, \beta)$. However, the function $\Gamma^{-1}$ is clearly discontinuous and so, in order to incorporate it into the bilevel optimisation model in \eqref{eqn:solution_map}--\eqref{eqn:Upper Level}, we approximate \eqref{eqn:StepFunction} with the parameterised hyperbolic tangent function:
\begin{equation*}
    t(v, \alpha, \beta) := \frac{\tanh(\alpha(v - \beta)) + 1}{2},
\end{equation*}
where $\alpha$ is a parameter that defines how steep the slope is, as is illustrated in Figure \ref{fig:VaryingAlpha}. Changing the value of $\beta$ will move the slope and hence change the proportion of the feature's values that are sampled, as illustrated in Figure \ref{fig:VaryingBeta}. By giving the adversary control over these parameters, we allow them to control the sampling of this feature.

For simplicity of notation, we collect parameters $\alpha$ and $\beta$ into the following block variable
\[
\theta:=(\alpha,\, \beta).
\]
Then, given some pre-defined $z \in (0,1)^{m \times q}$, we take a single row $z_i = (z_{i1}, \dots z_{iq}), \, i \in \{1\dots m\}$ where each $z_{ij} \sim U(0,1) \; \forall \; j \in \{1, \ldots, q\}$, and sample a single row of adversarial by
\begin{equation}
\label{eqn:Generator}
    G(z_i, \theta) := \begin{pmatrix}
        t(z_{i1}, \alpha_1, \beta_1) \\
        \vdots \\
        t(z_{iq}, \alpha_q, \beta_q)
    \end{pmatrix}^T.
\end{equation}
This generator function can now be directly implemented into our bilevel program \eqref{eqn:Upper Level} to model the adversary as a stochastic player who samples data from the distribution parameterised by $\theta$.

\section{Solving the training problem under protection}
\label{sctn:Solution Method}
As we have shown above, bilevel optimisation provides an effective way to model the interaction between our two players (i.e., the learner and the adversary), with the freedom to make their optimal decisions based on the defined objective functions. 
However, such a model is quite complicated to solve.
Our aim in this section is to propose a tractable and efficient method to solve the pessimistic bilevel optimisation problem in \eqref{eqn:solution_map}--\eqref{eqn:Upper Level}. Before we proceed to present our method, we first make two key observations that motivate the construction of our solution algorithm. 

First, we show that our lower-level problem does not have a unique optimal solution as imposed in a similar context in the paper by \cite{Brückner_Scheffer_2011}. 
More precisely, in the following proposition, we demonstrate by a simple example where the adversary only generates a single sample of data, that the set of optimal solutions to the lower-level problem, i.e., the $S(w)$ in \eqref{eqn:solution_map}, can indeed posses multiple values. 

\begin{proposition}
\label{prop:multipleSolutionns}
Let $w \in \mathbb{R}^q$ be such that $S(w)$ defined as in \eqref{eqn:solution_map} is non-empty. Suppose the adversary generates only one sample of data, $z \in (0,1)^{1 \times q}$.  
Then the lower-level problem defined by \eqref{eqn:solution_map} admits multiple optimal solutions.

\end{proposition}
\proof{Proof}
    Let $\theta^* := (\alpha^*, \beta^*) \in S(w)$ with $\alpha^*, \beta^* \neq 0$. Define $\theta^\prime = (-\alpha^*, 2z - \beta^*)$, and observe that for any $j \in \{1,\dots,q\}$, we have 
    \begin{align*}
        t(z_{j}, -\alpha_j^*, 2z_{j} - \beta_j^*) & = \frac{\text{tanh}(-\alpha_j^*(z_{j} - (2z_{j} - \beta_j^*))) + 1}{2} \\
        & = \frac{\text{tanh}(-\alpha_j^*(\beta_j^* - z_{j})) + 1}{2} \\
        & = \frac{\text{tanh}(\alpha_j^*(z_{j} - \beta_j^*)) + 1}{2} \\
        & = t(z_{j};\alpha_j^*, \beta_j^*).
    \end{align*}
    Therefore, $G(z, \theta^\prime) = G(z, \theta^*) \implies f(w, \theta^\prime) = f(w, \theta^*) \implies \theta^\prime \in S(w)$. \hfill \halmos
\endproof
    

Secondly, most classical approaches to solve continuous bilevel optimisation problems require the lower-level problem \eqref{eqn:solution_map} to be convex; see, e,g., \cite{Found_Bilevel_Prog, Bilevel_Optimisation_Adv_and_Chal} and references therein. Unfortunately, this is also not possible in the context of our problem,  given that the lower-level objective function $f(w, \theta)$ in \eqref{eqn:solution_map}, with respect to the lower-level variable $\theta = (\alpha, \beta)$, is not convex. We show this in the following proposition through a simple example where the adversary only generates a single sample of data and where the lower-level (adversary's) objective function is defined to be the logistic loss with opposite class labels. The idea being that the adversary will attempt to generate data that appears to the learner as the opposite of its true class, for example, they might wish to generate spam emails that appear legitimate. Note that the logistic loss has been shown to be a consistently high performing choice by \cite{Brückner_Scheffer_2011}, for example.

\begin{figure}[]
     \FIGURE
     {\includegraphics[scale=0.85]{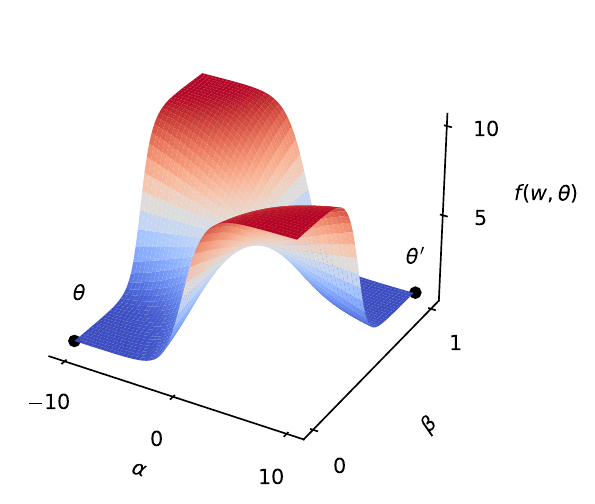}} 
{{\textmd{Illustration of the nonconvexity of the lower--level objective function\\ $f(w, \theta)$ with respect to $\theta$ where $w=10$, $z=0.5$, $\gamma = 1$ and $\mu = 0.1$.}} \label{fig:ConvexityPlot}}
{
}
\end{figure}

\begin{proposition}\label{Prop:Nonconvexity}
    Suppose the adversary generates only a single sample of adversarial data, $z \in (0,1)^{1 \times q}, \gamma \in \{0,1\}$. Let $f(w, \theta)$ be defined as in \eqref{eqn:lower}, where
    \[
    \ell(w, G(z,\theta), \gamma) := (\gamma - 1) \log(\sigma(w,G(z, \theta))) - \gamma \log(1 - \sigma(w, G(z, \theta))) + \mu||w||^2_2
    \]
with $w \in \mathbb{R}$,  $\mu \in \mathbb{R}$ being a regularisation parameter and $\sigma(w,G(z, \theta))$ given by
    \[
    \sigma(w,G(z, \theta)) := \frac{1}{1+e^{-w^TG(z, \theta)}}.
    \]
    Then the function $f(w, \theta)$ is not convex w.r.t. $\theta$.
\end{proposition}
\proof{Proof}
    Assume, without loss of generality, that $\gamma = 1, w>0$ and let $\xi \in (0,1)$ with $\xi \neq \frac{1}{2}$. Furthermore, let $\theta = (\alpha, \beta)^T \in \mathbb{R}^2$ be such that $\alpha(z - \beta) < 0$. Then consider $\theta^\prime = (-\alpha^\prime, 2z - \beta^\prime)$ and note from the proof of Proposition \ref{prop:multipleSolutionns} that $f(w, \theta^\prime) = f(w, \theta)$. It follows that 
    \[
    \xi f(w, \theta) + (1-\xi) f(w, \theta^\prime) = \xi f(w, \theta) + (1-\xi) f(w, \theta) = f(w, \theta).
    \]
  Subsequently, observe that
    \[\xi \alpha + (1-\xi) \alpha^\prime = \xi \alpha - (1-\xi) \alpha = (2 \xi - 1) \alpha,\]
    and similarly that
    \[z - (\xi \beta + (1-\xi)\beta^\prime) = z - (\xi \beta + (1-\xi)(2z - \beta)) = (2\xi - 1)(z - \beta).\]
    Therefore, we have
    \begin{align*}
        t(z, \xi \theta_i + (1-\xi) \theta_i^\prime) & = \frac{\text{tanh}\left ( (2\xi - 1)^2 \alpha_i(z - \beta_i) \right ) + 1}{2}, \; i = 1,\dots,q.
    \end{align*}
    Since $0 < (2\xi - 1)^2 < 1$, it holds that $(2\xi - 1)^2 \alpha (z - \beta) > \alpha (z - \beta)$. Define
    \[\bar{t}(u) := \frac{\text{tanh}(u) + 1}{2},\]
    for any $u \in \mathbb{R}$ and observe that
    \[\frac{\partial \bar{t}}{\partial u} (u) = \frac{1 - \text{tanh}^2(u)}{2} > 0,\]
    since $-1 < \text{tanh}(u) < 1$ and so $\text{tanh}^2(u) < 1$. Note also that for any $v \in \mathbb{R}^q$, we have that
    \[\frac{\partial \sigma}{\partial v_i}(w, v) = w_i\sigma(w,v)(1 - \sigma(w,v)) > 0, \; i = 1,\dots,q,\]
    since $0 < \sigma(w, v) < 1$. Hence, $\bar{t}(x)$ and $\sigma(w, x)$ are both strictly increasing. Since the log function is also strictly increasing, it follows that,
    \begin{align*}
        (2\xi - 1)^2 \alpha (z - \beta) > \alpha (z - \beta) & \implies t(z, \xi \theta + (1-\xi) \theta^\prime) > t(z; \theta) \\
        & \implies \sigma(w, t(z; \xi \theta + (1-\xi) \theta^\prime)) > \sigma(w, t(z; \theta)) \\
        & \implies f(w, \xi \theta + (1-\xi) \theta^\prime) > f(w, \theta) \\
        & \implies f(w, \xi \theta + (1-\xi) \theta^\prime) > \xi f(w, \theta) + (1 - \xi) f(\theta^\prime).
    \end{align*}
 Hence, the function $f(w, \theta)$ does not satisfy the convexity property w.r.t. $\theta$. 
 \hfill \halmos
\endproof
Note that Figure \ref{fig:ConvexityPlot} illustrates both propositions above, as for $w=10$, we have $f(w, \theta) = f(w, \theta')$ (Proposition \ref{prop:multipleSolutionns}) and the function $f(w, \theta)$ is clearly nonconvex (Proposition \ref{Prop:Nonconvexity}). 

In the solution approach that we propose here for problem \eqref{eqn:solution_map}--\eqref{eqn:Upper Level}, we neither require that the lower-level problem has a unique optimal solution, nor that it is convex. We assume that the loss functions are twice continuously differentiable and suppose that $z \in \mathbb{R}^{m \times q}$ has been determined prior to solving the bilevel optimisation problem. Considering the nonconvexity of the lower-level problem, the first step in our approach is to use the so-called value function reformulation 
\begin{equation}\label{eqn:solution_map_LLVF}
    S(w) = \left\{\theta \in \mathbb{R}^{2q} \mid f(w,\theta)-\varphi(w)\leq 0\right\},
\end{equation}
for our lower-level problem \eqref{eqn:solution_map}, 
where, indeed, $\varphi$ denotes the optimal value function of our lower-level optimisation problem; i.e.,  
\begin{equation}\label{eqn:solution_map_phi}
    \varphi(w) := \min_{\theta \in \mathbb{R}^{2q}} f(w,\theta).
\end{equation}

Secondly, we use the concept of two-level value function introduced in the papers \cite{doi:10.1137/110845197,dempe2014necessary}, 
to construct necessary optimality conditions that capture the structure of problem \eqref{eqn:solution_map}--\eqref{eqn:Upper Level}, and serve as base for our numerical algorithm. Note that the \textit{two-level (optimal) value function} associated to the optimisation problem \eqref{eqn:solution_map}--\eqref{eqn:Upper Level}  is defined as 
\begin{equation}\label{TwoLevelValue}
    \varphi_p(w) := \max_{\theta \in S(w)} F(w,\theta),
\end{equation}
and is called as such given that it involves two-levels of decision-making (i.e., the \textit{max} operator at the upper-level for the leader's objective function and the \textit{min} operator in the lower-level for the follower's objective function). 
Based on this concept, a point $w$ will be said to be a local optimal solution of problem \eqref{eqn:solution_map}--\eqref{eqn:Upper Level}  if there is a neighbourhood  $W$ of $w$ such that 
\[
\varphi_p(w)\leq \varphi_p(w') \;\, \mbox{ for all } w' \in W.
\]
Next, note that one of the classical approaches to solve a continuous optimisation problem is to directly compute its stationary points; in other words, this corresponds to computing points that satisfy necessary optimality conditions of the corresponding problem (see, e.g., \cite{fischer1992special} and references therein). To proceed with this type of approach for problem \eqref{eqn:solution_map}--\eqref{eqn:Upper Level}, we first need to construct its necessary optimality conditions, and this is what we do in the next result.
\begin{theorem}\label{Thm:Necessary Optimality}If $w$ is a local optimal solution of problem \eqref{eqn:Upper Level}, where we suppose that the set   gph$S:=\left\{(w', \theta')|\; \theta'\in S(w')\right\}$, with $S$ defined in \eqref{eqn:solution_map}, is closed, there exists $\lambda \geq 0$ such that
\begin{align}\label{eqn:system}
    \nabla_w F(w, \theta)=0,\\
    \nabla_{\theta} F(w, \theta) - \lambda \nabla_{\theta} f(w, \theta)=0, \label{eqn:system-2}
\end{align}
where $(w,\theta)$ is such that $\theta\in S_p(w):=\arg\underset{\theta'}{\max}\left\{F(w, \theta')\left|\;\, \theta'\in S(w)\right.\right\}$ and satisfies the assumptions: 
\begin{itemize}
    \item[(i)] For every sequence $w_k\rightarrow w$, there is a sequence of $\theta_k\in S_p(w_k)$ such that that $\theta_k \rightarrow \theta$.  
    \item[(ii)] There exists $(\rho, \delta)\in \mathbb{R}^2_+$ such that the following condition holds:  
    \[
    d(\theta', \, S(w'))\leq \rho (f(w', \theta') - \varphi(w')) \;\; \mbox{ for all }\;\; (w', \theta')\in \mathbb{B}((w, \theta), \delta).
    \]
Here, $d(\theta', S(w'))$ denoting the distance from the point $\theta'$ to the set $S(w')$. 
\end{itemize}
\end{theorem}
\proof{Proof}
Start by noting that considering the expression \eqref{TwoLevelValue}, problem \eqref{eqn:solution_map}--\eqref{eqn:Upper Level} can be written as 
\[
\min_{w' \in \mathbb{R}^{q}} \varphi_p(w').
\]
Hence, the first step of the proof consists of showing that the two-level value function $\varphi_p(w')$ is Lipschitz continuous near $w$. Considering  the continuous differentiability of the upper-level objective function $F(w', \theta')$ \eqref{eqn:UpperLevel} and the inner semicontinuity of the set-valued mapping $S_p(w')$ at $(\theta, w)$ as assumed in (i), it follows from \cite{doi:10.1137/110845197}, for example, that it suffices to show that the set-valued mapping $S(w')$ is Lipschitz-like in the sense of \cite{d34a10f6-7b74-3493-a576-5928330b333c} around the point $(w, \theta)$,
meaning that there exist neighbourhoods $W$ of $w$, $\Theta$ of $\theta$ and a constant $l > 0$ such that
\[d(\theta', S(w')) \leq l ||v' - w'|| \;\, \mbox{for all} \;\, v',w' \in W \text{ and } \theta' \in S(v') \cap \Theta,\]
where $d$ is a function measuring the distance form a point to a subset of $\mathbb{R}^{2q}$.

Since $S_p(w') \subseteq S(w')$ for all $w'\in \mathbb{R}^q$, the condition assumed in (i), for the set-valued mapping $S_p(w')$,  also holds at the point $(w, \theta)$ for  $S(w')$ defined in \eqref{eqn:solution_map_LLVF}. Furthermore, note that since $w\in S(\theta)$, we have $\nabla_{\theta} f(w,\theta)=0$. Hence, the lower-level optimal value function $\varphi(w')$ is strictly differentiable at $w$, as its Clarke subdifferential reduces to just  $\nabla_w f(w, \theta)$. This means that the function $(w', \theta') \longmapsto f(w',\theta')-\varphi(w')$ defining $S(w')$ in \eqref{eqn:solution_map_LLVF} is strictly differentiable at $(w, \theta)$. 

It therefore follows that, for any $\theta^*\in \mathbb{R}^{2q}$, we can find some number $\lambda \geq 0$ such that the \textit{coderivative} (i.e., the set-valued mapping generalised derivative in the sense of Mordukhovich \cite{mordukhovich2018variational}) of $S(w')$ \eqref{eqn:solution_map_LLVF} is obtained at the point $(w, \theta)$ as
\begin{equation}\label{eqn:DS}
    D^*S(w|\theta)(\theta^*)=\left\{\begin{array}{cl}
    \{0\} & \mbox{ if } \;\theta^*=-\lambda \nabla_{\theta} f(w, \theta), \\
    \emptyset & \mbox{ otherwise}.
\end{array} \right.
\end{equation}
In fact, as gph$S$ is closed and assumption (ii) is satisfied at $(w, \theta)$, it holds that 
\[
\begin{array}{lll}
 D^*S(w|\theta)(\theta^*) &:=&\left\{w^*\in \mathbb{R}^q\left|(w^*, -\theta^*)\in N_{\text{gph}S}(w, \theta)\right.\right\}\\[2ex]
                 &\subseteq & \left\{w^*\in \mathbb{R}^q\left|\, \left(w^*, -\theta^*\right)\in \left\{\lambda \left.\left[\begin{array}{c}
                      \nabla_{w} f(w, \theta) - \partial \varphi(w)\\
                      \nabla_{\theta} f(w,\theta)
                 \end{array}\right]\right|\lambda \geq 0\right\}\right.\right\}
\end{array}
\]
given that $f(w, \theta)-\varphi(w)=0$ and $\partial \varphi(w)=\left\{\nabla_w f(w, \theta)\right\}$ with the inclusion following from \cite{doi:10.1137/S1052623401395553}.
{\color{black}{Here, $\partial$ represents the Clarke subdifferential concept, while for a closed set $C\subset \mathbb{R}^n$, $N_{C}(\bar x)$ corresponds to the limiting normal cone to $C\subset \mathbb{R}^n$ at the point $\bar{x}\in C$, defined by 
\begin{equation}\label{basic normal cone}
 N_{C}(\bar x):= \left\{v\in \mathbb{R}^n\left|\; \exists v_k \rightarrow v, \, x_k \rightarrow \bar x \,(x_k\in C): \, v_k\in \widehat{N}_{C}(x_k)\right.\right\},
\end{equation}
where $\widehat{N}_{C}$ denotes the regular normal cone to $C$ (see, e.g., \cite{mordukhovich2018variational} for more details):
\[
 \widehat{N}_{C}(\bar x):=\left\{v\in \mathbb{R}^n \left |\; \langle v, u-\bar x\rangle \leq o(\|u-\bar x\|)\;\; \forall u\in C\right. \right\}.
\]

It therefore follows from \eqref{eqn:DS} that $D^*S(w|\theta)(0)=\{0\}$ given that $\nabla_{\theta} f(w,\theta)=0$ considering the fact that $\theta\in S(w)$. Hence, the set-valued mapping $S$ is Lipschitz-like around $(w, \theta)$. Combining this with assumption (i), it follows that $\varphi_p$ is Lipschitz continuous around $w$ and we have
\begin{equation}\label{eqn:MainStepforOpt}
    \left\{
\begin{array}{l}
 \partial \varphi_p(w) =-\partial \varphi_{op}(w) \subseteq \nabla_w F(w, \theta) - D^*S(w, \theta)\left(-\nabla_{\theta} F(w, \theta)\right) = \left\{\nabla_w F(w, \theta)\right\}\\[1ex]
 \mbox{with }\;\; \nabla_{\theta} F(w, \theta) -\lambda \nabla_{\theta} f(w, \theta)=0 \;\;\mbox{ for some }\;\; \lambda \geq  0\\[1ex]
\mbox{and }\quad \varphi_{op}(w'):=\underset{\theta'}{\min}\left\{-F(w', \theta')\left|\;\, \theta'\in S(w')\right.\right\},
\end{array}
\right.
\end{equation}
while considering equation \eqref{eqn:DS}; cf. \cite{dempe2014necessary}. Subsequently, recalling that $0\in \varphi_p(w)$ since $w$ is a local optimal solution of problem \eqref{eqn:Upper Level}, condition \eqref{eqn:MainStepforOpt} leads to the result; i.e., the fulfilment of the system of equations in \eqref{eqn:system} and \eqref{eqn:system-2}. \hfill \halmos
\endproof

Before a deep analysis of the result, note that the system \eqref{eqn:system}--\eqref{eqn:system-2} corresponds to necessary conditions for the point $(w, \theta)$ to be a saddle point (or Nash equilibrium) for the special class of minmax problem described by the pessimistic bilevel optimisation problem \eqref{eqn:Upper Level}, \eqref{eqn:solution_map_LLVF}.  Recall that $(w, \theta)$ is a saddle point (or Nash equilibrium) for this minmax problem if and only if 
\[
\underset{w' \in \mathbb{R}^q}\min~F(w',\theta) = F(w,\theta) = \underset{\theta' \in S(w)}\max~F(w, \theta').
\]
Note however that we are not in a convex-concave situation as it is common in the minmax programming literature; see, e.g., \cite{jin2020local} and references therein on the existence of saddle points, and so the requirement to develop more advanced techniques.

Moreover, note that assumption (i) in the theorem is a very well-known concept in variational analysis called inner semicontinuity, which can automatically hold for $S_p$ at $(w, \theta)$ if $S_p(w)=\{\theta\}$; it is closely related to the better-known, but stronger concept of lower semicontinuity. Our result can also hold if the weaker inner semicompactness, which holds at a point $w$, in particular, if $S_p$ is just uniformly bounded; i.e., if there exists a neighbourhood $W$ of $w$ and a bounded set $B$ such that $S_p(w) \subset B$ for all $w\in W$.  For more details on condition (i), see, e.g., \cite{doi:10.1137/110845197, mordukhovich2018variational,dempe2014necessary} and references therein. 

Given that for assumption (ii) in the theorem, it corresponds to the uniform weak sharp minimum condition associated to our lower-level problem; for a detailed discussion of this assumption in the context of the lower-level problem \eqref{eqn:solution_map_LLVF} and two-level value function \eqref{TwoLevelValue}, interested readers are referred to Section 5 of \cite{doi:10.1137/110845197} and references therein.

As our aim here is to compute points satisfying the necessary optimality conditions in \eqref{eqn:system}--\eqref{eqn:system-2}, a key question to ask here is how rich is this system in terms of capturing the inclusions $y\in S_p(w)$ and $y\in S(w)$, which are crucial in the definition of our pessimistic bilevel optimisation problem, given in \eqref{eqn:Upper Level}, \eqref{eqn:solution_map_LLVF}. Obviously, inclusion $S_p(w)\subset S(w)$ holds by definition. But we can check that, under suitable assumptions, condition \eqref{eqn:system-2} only represents a necessary condition for $\theta \in S_p(w)$ under the lower-level value function formulation $S$ in \eqref{eqn:solution_map_LLVF}, given that the inner problem 
\[
\max_{\theta \in S(w)} F(w,\theta)
\]
of the minmax program \eqref{eqn:Upper Level} is nonconcave as $S(w)$ is nonconvex. With this in mind, it is not guaranteed that lower-level optimality is automatically ensured in the system \eqref{eqn:system}--\eqref{eqn:system-2}. Hence, to enrich this system, we append the necessary condition $\nabla_{\theta} f(w,\theta)=0$ for $\theta\in S(w)$ to it.

Secondly, to incorporate the non-negativity, we replace $\lambda$ with $\zeta^2$, which brings us to solving
\begin{equation}\label{eqn:system-12}
     \Phi(w, \theta, \zeta) := \begin{pmatrix}
    \nabla_w F(w, \theta) \\
 \nabla_{\theta} F(w, \theta) - \zeta^2 \nabla_{\theta} f(w, \theta) \\
\nabla_{\theta} f(w, \theta)
\end{pmatrix} =0,
\end{equation}
where $\zeta$ is a real number and new variable of the system.

\begin{algorithm}[]
\caption{Method for classification under strategic adversary
Manipulation}
\label{alg:LVM}
\begin{algorithmic}[1]
\Require Function $\Phi$, starting points $w^0, \theta^0$, $\eta^0$ a tolerance $\epsilon$, 
maximum number of iterations $\text{max\_iter}$ and parameters and $\kappa > 0, \tau \in [0,1]$
\While{$||\Phi(w^k, \theta^k)||^2 < \epsilon$ and $k < \text{max\_iter}$}
    \State solve $(\nabla \Phi(w^k, \theta^k)^T \nabla \Phi(w^k, \theta^k) + \eta I)d^k = - \nabla \Phi(w^k, \theta^k) ^T \Phi(w^k, \theta^k)$
    \While{$||\Phi((w^k, \theta^k) + \omega^k d^k)||^2 \geq ||\Phi(w^k, \theta^k)||^2 + \omega^k \nabla \Phi(w^k, \theta^k)^T \Phi(w^k, \theta^k) d^k$}
        \State $\omega \gets \frac{\omega}{2}$
        \If{$\omega < 10^{-100}$}
        \State set $\eta := \eta^0$
        \State \textbf{break}
    \EndIf
    \EndWhile
    \If{$\frac{||\Phi(w^k, \theta^k)||^2}{||\Phi(w^{k-\kappa}, \theta^{k-\kappa})||} > \tau$}
        \State $\eta \gets \frac{\eta}{10}$
    \EndIf
    \State set $(w^{k+1}, \theta^{k+1}) := (w^k, \theta^k) + \alpha d$
    \State set $k := k + 1$
\EndWhile
\end{algorithmic}
\end{algorithm}

Since the system of equations \eqref{eqn:system-12} is overdetermined with $3q+1$ variables and $5q$ equations, one of the most tractable approaches to solve it is the well-known Levenberg-Marquardt (LM) method (e.g., \cite{ahookhosh2022finding,jolaoso2024fresh}), which we summarise in Algorithm~\ref{alg:LVM}. Clearly, to implement this algorithm requires the Jacobian of $\Phi(w, \theta, \zeta)$, which is given by
\begin{equation}
\label{eqn:jacobian}
    \nabla \Phi(w, \theta, \zeta) = \begin{pmatrix}
    \nabla^2_{w w} F(w,\theta) & \nabla^2_{w \theta} F(w,\theta) & 0 \\
    \nabla^2_{w \theta} F(w,\theta)  - \zeta^2 \nabla^2_{w \theta} f(w,\theta) & \;\nabla^2_{\theta \theta} F(w,\theta) - \zeta^2 \nabla^2_{\theta \theta} f(w,\theta) & \;- 2 \zeta \nabla_{\theta} f(w,\theta) \\
    \nabla^2_{w \theta} f(w \theta) & \nabla^2_{\theta \theta} f(w,\theta) & 0 \\
\end{pmatrix}.
\end{equation}
The precise formulas of the derivatives in \eqref{eqn:system-12} and \eqref{eqn:jacobian} under the logistic loss, which we use in the experiments in section \ref{sctn:Applications}, can be found in the Appendix.

The LM method is closely related to the Gauss-Newton method, with the exception that the LM method includes a perturbation on the term $\nabla \Phi(w^k, \theta^k)^T \nabla \Phi(w^k, \theta^k)$, controlled by the parameter $\eta$, which ensures the term is invertible. When $\eta = 0$, the LM method coincides with the Gauss-Newton method. The value of $\eta$ is decreased as the algorithm approaches the solution of equation \eqref{eqn:system-12}; i.e., to a stationary point of our pessimistic bilevel optimization problem \eqref{eqn:Upper Level}. There are numerous approaches to selecting the best $\eta$ at each iteration, such as selecting its value proportionally to $||\Phi(w, \theta)||^2$. Since we found this method to perform inconsistently, we opt for decreasing $\eta$ when the objective value shows slow progress. As shown in step 6 of Algorithm~\ref{alg:LVM}, we compare the value of $||\Phi(w, \theta)||^2$ at the current iteration to that of its previous value some number of iterations, $\kappa$, prior.
If the ratio of these two exceeds some constant $\tau$, we decrease $\eta$ by a factor of 10. We found the values $\kappa = 5, \tau = 0.9$ to perform consistently. Steps 3--8 of the algorithm perform a line search to find a sufficient value of $\omega$. In the case that the line search cannot find a suitable value of $\omega$ satisfying the condition in step 3, then we reset $\eta$ and break from the loop, as shown in steps 5--8.

\section{Applications and numerical experiments}\label{sctn:Applications}
In this section, we assess the performance of our model's ability to classify unseen data from the future with two experiments.
In the first, we construct the \textit{TREC} dataset as the compilation of the email corpora provided  for the NIST Text Retrieval Conferences, \citep{trec06, trec07}. These are collections of spam and legitimate emails received throughout the years of 1993--2007, which when compiled in a dataset contain a total of 68338 emails.
The second experiment makes use of the \textit{Amazon} dataset, a collection of 3185845 fake and legitimate reviews for cell phones and their accessories made between the years of 1998--2014, collected by \cite{amazon}.
Both datasets are organised chronologically with the earliest 2000 samples taken as the training set. The remaining data are separated by year to form multiple test sets. In both cases, the data are tokenised and, after removing the stop words, the occurrence of each word in the training set is counted. We take the top 1000 most commonly occurring words and convert the data into binary bag-of-word vectors.
In the experiments, we only allow the adversary to generate data with the adversarial-class label: in the \textit{TREC} spam email experiment we only allow the adversary to generate spam emails and in the \textit{Amazon} fake reviews experiment we only allow the adversary to generate fake reviews.
The amount of data the adversary can generate is controlled by the constant $\rho$. We investigate setting $\rho$ to various proportions of the size of the adversarial-class data in the training set, including $5\%, 7.5\%, 10\%, 12.5\%, 15\%, 20\%, 25\%$.

For all the experiments, we define the learner's prediction function as
\begin{equation*}
    \sigma(w,X_i) := \frac{1}{1+e^{-w^TX_i}}, \; i = 1,\dots,n,
\end{equation*}
which gives the probability that $X_i$ belongs to class $1$.
The learner's loss function is defined as the logistic loss with $l_2$ norm,
\[\mathcal{L}(w, X_i,y_i) := -y_i \log(\sigma(w,X_i)) - (1 - y_i) \log(1 - \sigma(w,X_i)) + \mu||w||^2_2, \; i = 1,\dots,n,\]
where $y_i$ is the true class of $X_i$ and $\mu \in \mathbb{R}$ is a regularisation parameter of which we consider a selection of values in the range $\mu \in [10^{-6}, 10^2]$. The logistic loss function is also used for the adversary, but with the class labels swapped to their opposite value; i.e., 
\[
\ell(w, G(z_i,\theta), \gamma_i) := \mathcal{L} (w, G(z_i,\theta), 1 - \gamma_i), \; i = 1,\dots,m
\]
for class labels $\gamma_i \in \{0,1\}$, and for all $i = 1,\dots,m$, this formula expands to
\[\ell(w, G(z_i,\theta), \gamma_i) = (\gamma_i - 1) \log(\sigma(w,G(z_i, \theta))) - \gamma_i \log(1 - \sigma(w, G(z_i, \theta))) + \mu||w||^2_2.\]
The derivatives formulas needed to implement the system in \eqref{eqn:system-12} and the Jacobian in \eqref{eqn:jacobian}, in the context of Algorithm \ref{alg:LVM},   can be found in the Appendix.
\begin{figure}[]
     \FIGURE
     {\includegraphics[scale=0.45]{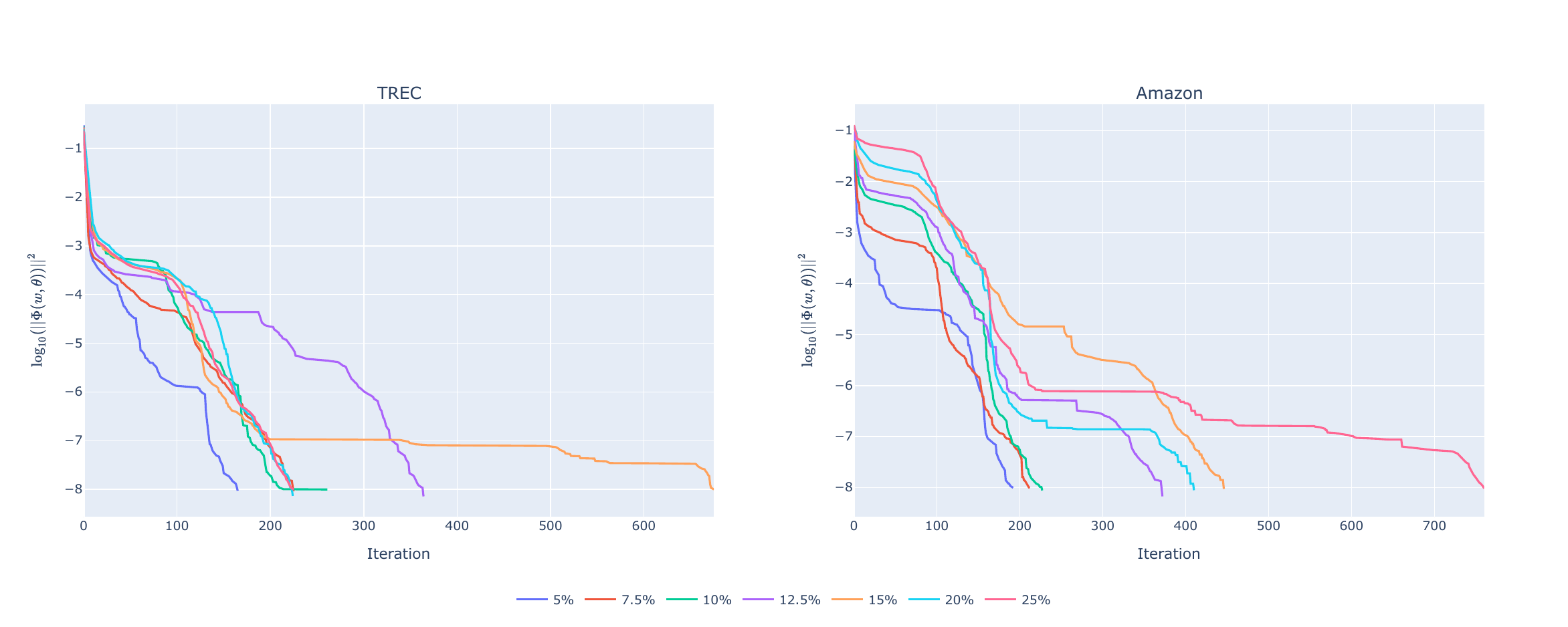}} 
{{\textmd{LM convergence}} \label{fig:LVM Convergence}}
{
}
\end{figure}

We solve  \eqref{eqn:system-12} by implementing the Levenberg-Marquardt (LM) method in Algorithm \ref{alg:LVM}. We set $\text{max\_iter} = 1000$ and set the stopping tolerance to $\epsilon = 10^{-8}$. We set the start points $w_i^0 = 0, \alpha_i^0 = 1000 \, \forall \, i = 1,\dots,q$. The experimental results showed the model to be highly sensitive to the start point of $\beta$. While many start points should be tested, we found the most consistent method to be starting $\beta$ such as to best replicate a sample of the training data. In this way, the adversary can use the generator to modify existing data rather than generating completely new samples. We discuss this strategy further later in this section. We initialise $\eta = 0.001$ and decrease its value by comparing the function value at the current iteration to that of its value 5 iterations prior ($\kappa = 5$), and decrease $\eta$ by a factor of 10 when the ratio between the function values is greater than $\tau = 0.9$.
To investigate the convergence of the LM algorithm, we record the logarithm with base 10 of the sum of squares of the system for a selection of experiments in Figure \ref{fig:LVM Convergence} to illustrate how the amount of adversarial data impacts the convergence rate. It is clear from these plots that the LM method is effective in solving system \eqref{eqn:system-12} since the algorithm converged successfully to the desired tolerance in all cases. While the \textit{TREC} experiments do not show a clear pattern relating $\rho$ with convergence rate, the \textit{Amazon} experiments suggest that higher values of $\rho$ correspond to a slower convergence rate.

Adversarial datasets are typically imbalanced, with the proportion changing with time. While the F1
score is a popular choice for imbalanced datasets, its asymmetry can lead to misleading results. When a dataset has a positive class majority, a classifier that accurately detects the positive class (e.g. spam emails) will record a high F1 score even if its ability to accurately detect the negative class (e.g. legitimate emails) is poor, making it difficult to identify which classifier performs best. For example, an email filter might play it safe by simply classifying every email as spam. On a spam-majority test set, this classifier will record a high F1 score while in reality providing no meaningful classifications. Therefore, we instead measure a classifier's performance by the symmetric $P_4$ metric \citep{p4_metric}, which is defined by
\[P_4 := \frac{4 \cdot \text{TP} \cdot \text{TN}}{4 \cdot \text{TP} \cdot \text{TN} + (\text{TP} + \text{TN}) \cdot (\text{FP} + \text{FN})},\]
where TP, TN, FP and FN represent the counts of true positives, true negatives, false positives and false negatives, respectively. In this way, we gain a better understanding of a classifier's ability to discern between classes.
\begin{figure}[H]
     \FIGURE
     {\includegraphics[scale=0.45]{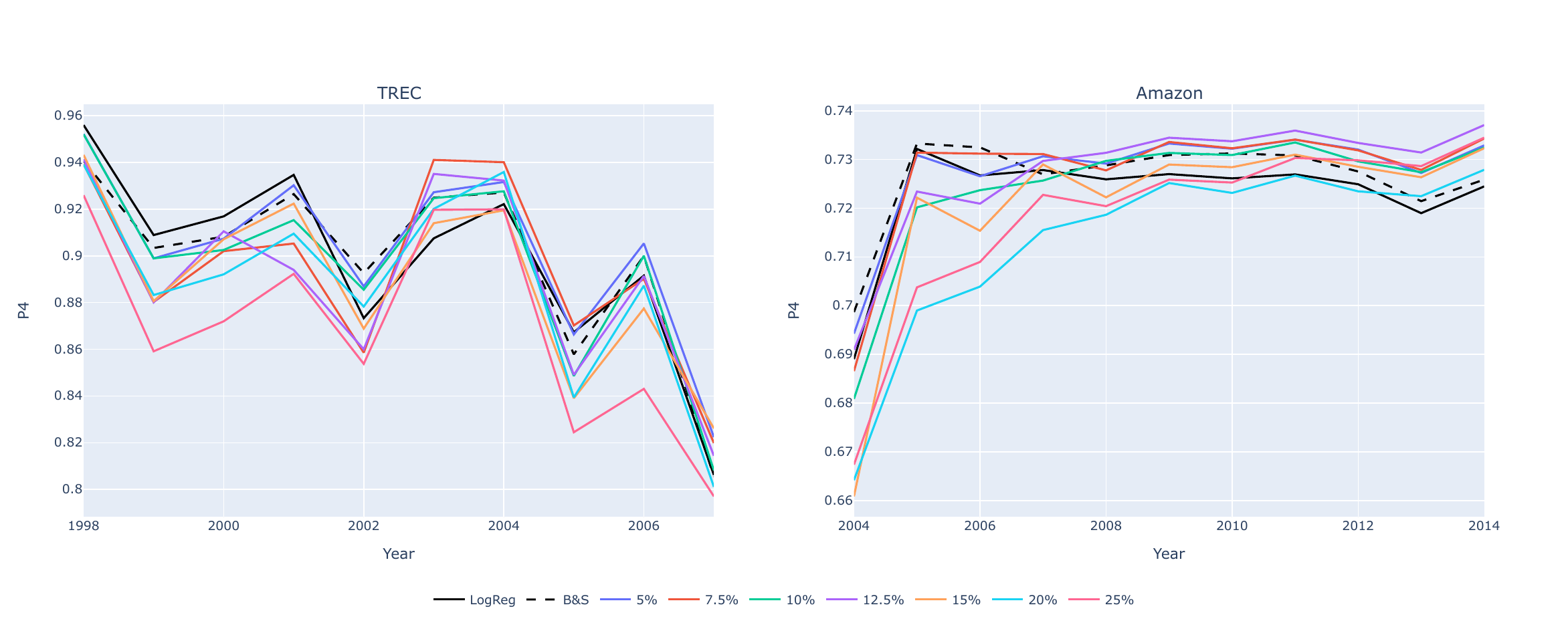}} 
{{\textmd{Performance comparison}} \label{fig:SpamPerformance}}
{
}
\end{figure}
To better investigate the affect of limiting the adversary's influence over the performance, we plot the $P_4$ score for the same selection of classifiers as in Figure \ref{fig:LVM Convergence} to illustrate how performance changes with $\rho$. These are plotted in Figure~\ref{fig:SpamPerformance} for both datasets. We also compare the performance to the standard logistic regression classifier without adversarial influence (\textit{LogReg}) and the bilevel optimisation model from \cite{Brückner_Scheffer_2011} (\textit{B\&S)}. As expected, the standard logistic regression model sees a steady decrease in performance with time on the spam email data in the \textit{TREC} experiments. Similarly, after an initially poor performance on the 2004 test set, we see a gradual decline in performance for the standard logistic regression classifier on the \textit{Amazon} data, although the change in performance is less so than on the \textit{TREC} data. This could suggest a weaker attempt by adversaries to modify their fake reviews compared to spam emails.

In the \textit{TREC} experiments, the earlier dates (up to the year 2001) see no noticeable improvement in performance by our bilevel optimisation model. In fact, we see a worse performance in general. This is to be expected, since the training data has been supplemented with adversarial data that likely does not follow the same distribution as the rest of the training data. Instead, it is in later years that we see noticeable improvements in performance suggesting the data generated in the bilevel optimisation model better represents the distribution from those years. A similar pattern is seen in the \textit{Amazon} experiment where the bilevel optimisation model sees a worse or matched performance in 2004 and 2005 but with a consistently higher performance in later years.

It is clear that the amount of data the adversary generates can impact the performance. In both experiments we see a noticeable variation in performance with $\rho$ for each year. The lower values of $\rho$ consistently outperform the others, with either $\rho = 5\%$ or $\rho = 7.5\%$ performing best. Conversely, we see the higher values of $\rho$ typically performing the worst. In particular, the \textit{TREC} experiment sees $\rho = 25\%$ consistently performing the worst for every year except 2003 and the \textit{Amazon} experiments seeing $\rho = 20\%$ perform worst for every year except 2004. These results suggest that higher values of $\rho$ give the adversary too much influence over the data, to the point that it no longer reflects reality. It is clear that a balance between the adversarial and static data should be investigated when training the model.

When comparing our model to the existing pessimistic model by \cite{Brückner_Scheffer_2011}, we see in the \textit{TREC} experiments that our model exceeds its performance for every year from 2003 for at least one value of $\rho$. In the \textit{Amazon} experiments, we see \textit{B\&S} outperform our model for all values of $\rho$ between 2004 - 2006 and then our model match or exceed \textit{B\&S} every year from 2007 for at least one value of $\rho$. In fact, between 2012 - 2014, we see our model outperform \textit{B\&S} for all values of $\rho$.

\begin{figure}[h]
\FIGURE
{
\subcaptionbox{Start points of $\beta$\label{fig:beta0}}
{\includegraphics[width=0.52\textwidth]{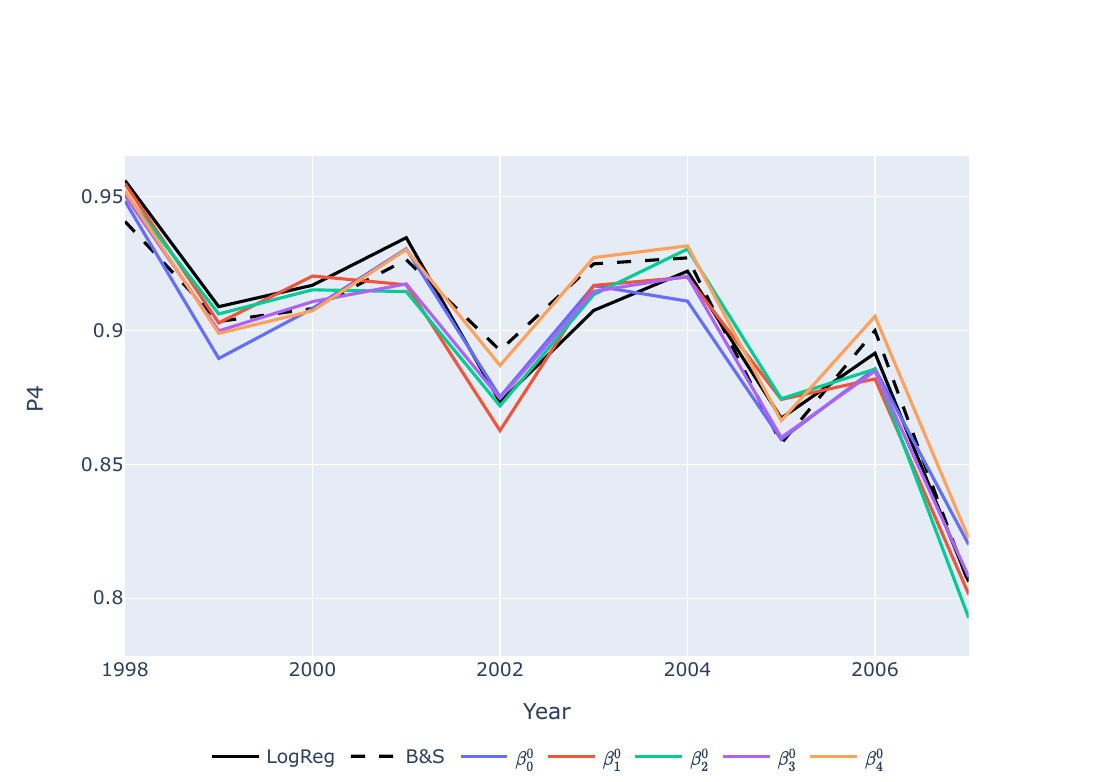}}
\subcaptionbox{Start points of $\zeta$\label{fig:zeta0}}
{\includegraphics[width=0.52\textwidth]{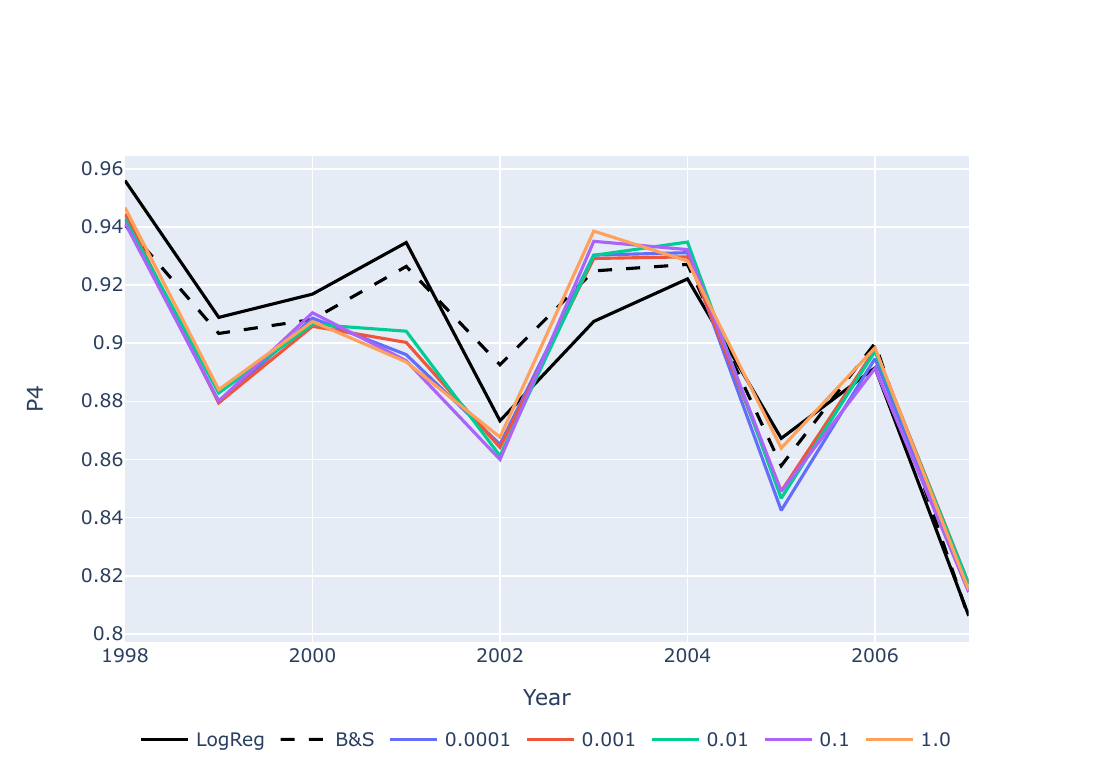}}
}
{
{\textmd{Varying the start points $\beta^0$ and $\zeta^0$}}
\label{fig:StartPointInvestigation}}
{
}
\end{figure}

Since we place no constraints on the parameters of the adversary's generator ($\alpha$ and $\beta$) the adversary has no restriction on the data they can generate. Throughout the experiments, we found that such an approach makes the model highly sensitive to the starting points of these parameters, in particular $\beta$, which controls the proportions of each feature in the adversary's data. We found the most consistent way of identify high-performing start points to be starting the adversary with parameters which generate data that closely resembles some already existing data. For binary bag-of-words static data $X$, we can achieve this by randomly sampling a subset of $b$ rows, $x \subset X, \, x \in \{0,1\}^{b \times q}$. Then, we set the start point of $\beta$ to rate of occurrence of each word in $x$, 
\[\beta^0_j = \frac{1}{b} \sum_{i=1}^b x_{ij}, \; j = 1,\dots,q.\]
In this way, we can generate a sample of data that closely follows the marginal distributions of the sample of static data. Now the adversary can modify this data to create adversarial data rather than generating new data from scratch. To illustrate the effect of the start point of $\beta$ on performance, Figure \ref{fig:beta0} shows an example of 5 randomly assigned subsets of the static data of the \textit{TREC} dataset where $\rho = 5\%$ and $\zeta^0 = 0.1$. Similarly, we found the start point of $\zeta$ to affect the performance. To investigate its effect, we grid search various powers of 10 and plot the performance in Figure \ref{fig:zeta0} where $\rho = 12.5\%$ and for a fixed start point of $\beta$. While there is no clear pattern, it is clear that various choices of this start point should also be tested.

It is clear from the experiments that under certain choices of the start point of $\beta$, the bilevel program proposed here can consistently predict an adversary's movements. While the initial accuracy on earlier dates sees some sacrifice, the bilevel program can provide more accurate predictions than a traditionally trained classifier on future data. Further to this, we also see improved performance over the existing approach of using pessimistic bilevel optimisation to model adversarial learning problems.

\section{Conclusion}\label{sctn:Conclusion}
In this work, we investigated modelling adversarial attacks at implementation time as a game between two players, the learner who trains a classifier and an adversary who generates data in an attempt to evade detection by that classifier. We formulated the game as a pessimistic bilevel optimisation problem with the learner taking on the role of the leader and the adversary as the follower. In this way, the adversary can observe the capabilities of the learner's classifier before deciding on the values of some parameterised distribution that they will draw new data from. Designing the adversary in this way allowed us to incorporate binary features into a continuous model and depart from the established literature which instead made use of a feature map. Unlike the use of feature maps, our approach allowed us to construct a continuous bilevel optimisation model that retains distinct multiple optimal solutions to the lower-level problem in their original feature space. While the existing solution method relies on assumptions about the convexity of the lower-level problem and the uniqueness of its solution, we constructed a method which makes no such assumptions and instead utilises the value function of the lower-level problem to construct a system of equations.


We demonstrated the performance of this solution method on two text-based adversarial datasets by measuring the resulting classifier's performance into the future. Where a typical classifier sees decreased performance over time, our model retained a consistently higher accuracy in later dates. We noted that since the adversary is modelled as an unrestricted data generator, the performance is highly sensitive to the start point of the parameters of the adversary's generator. However, by setting this start point such that the adversary begins with some existing data, we constructed a model that accurately anticipated the adversary's movements and trained resilient classifiers which saw improved performance over the existing pessimistic bilevel approach.


%
 
%
%



\bibliographystyle{informs2014} 






  



\begin{APPENDIX}{Leader and follower's derivatives under the logistic regression loss}
 Let $z \in (0,1)^{m \times q}$, i.e. we construct an adversary who generates $m$ samples of data. Let $\tilde{G}(z, \theta) \in \mathbb{R}^{m \times q}$ be the complete matrix of adversarial data, defined as
 \[\tilde{G}(z, \theta) := \begin{pmatrix}
     G(z_1, \theta) \\
     \vdots \\
     G(z_m, \theta)
 \end{pmatrix},\]
where for $k = 1,\dots m, \; G(z_k, \theta) \in \mathbb{R}^q$. The derivative of the upper-level objective function, $F$, with respect to the learner's weights is given by
\[\nabla_{w} F(w, \theta) = X^T(\sigma(w, X) - y) + \tilde{G}(z, \theta)^T(\sigma(w, \tilde{G}(z, \theta)) - \gamma).\]
Let $D_X = \text{diag}(\sigma(w, X))$ and $D_{G} = \text{diag}(\sigma(w, \tilde{G}(z, \theta)))$, where for some vector $v \in \mathbb{R}^n$, $\text{diag}(v)$ is the $n \times n$ diagonal matrix with diagonal elements $v$. The second derivative with respect to the learner's weights are then given by
\[\nabla^2_{ww} F(w, \theta) = \frac{1}{n} X D_X(I_n - D_X) X^T + \frac{1}{m} \tilde{G}(z, \theta) D_G(I_m - D_G) \tilde{G}(z, \theta)^T,\]
where $I_n$ and $I_m$ are the $n\times n$ and $m\times m$ identity matrices respectively. We define the derivatives with respect to the parameters of the adversary's generator element-wise. For a single sample (row) of adversarial data, given by $G(z_k, \theta), \, k=1,...,m$, we define a single feature of this sample as
\[G_i(z_{k}, \theta) := t(z_{ki}, \alpha_i, \beta_i), \, \mbox{ for } \, i = 1,\dots,q, \; k = 1,\dots,m,\]
and note the following derivatives:
\[
\begin{array}{rll}
 \frac{\partial t}{\partial \alpha_i}(z_{ki, \alpha_i, \beta_i}) & = & \frac{z_{ki} + \beta_i}{2 \text{cosh}^2(\alpha(z_{ki} + \beta_i))}, \, \mbox{ for } \, i=1, \ldots, q, \; k=1, \ldots, m,\\[2ex]
\frac{\partial t}{\partial \beta_i}(z_{ki}, \alpha_i, \beta_i)  & = & \frac{\alpha_i}{2 \text{cosh}^2(\alpha(z_{ki} + \beta_i))}, \, \mbox{ for } \, i=1, \ldots, q, \; k=1, \ldots, m,\\[2ex]
\frac{\partial^2 t}{\partial \alpha_i \partial\alpha_i}(z_{ki}, \alpha_i, \beta_i)  & = & \frac{-(z_{ki} + \beta_i)^2\text{tanh}(\alpha_i(z_{ki}+\beta_i))}{\text{cosh}^2(\alpha(z_{ki} + \beta_i))}, \, \mbox{ for } \, i=1, \ldots, q, \; k=1, \ldots, m,\\[2ex]
\frac{\partial^2 t}{\partial \beta_i \partial\beta_i}(z_{ki}, \alpha_i, \beta_i)  & = & \frac{-\alpha_i^2\text{tanh}(\alpha_i(z_{ki} + \beta_i))}{\text{cosh}^2(\alpha_i(z_{ki} + \beta_i))}, \, \mbox{ for } \, i=1, \ldots, q, \; k=1, \ldots, m,\\[2ex]
\frac{\partial^2 t}{\partial \alpha_i \partial\beta_i}(z_{ki}, \alpha_i, \beta_i) & = & \frac{1 - 2 \alpha_i (z_{ki} + \beta_i) \text{tanh}(\alpha_i(z_{ki} + \beta_i))}{2 \text{cosh}^2(\alpha_i(z_{ki} + \beta_i))} \, \mbox{ for } \, i=1, \ldots, q, \; k=1, \ldots, m.
\end{array}
\]
Then, the derivative of the upper-level objective function with respect to the parameters of the adversary's generator can be written as 
\[\frac{\partial F}{\partial \theta_i} (w, \theta) = \sum_{k=1}^m 
\frac{\partial\mathcal{L}}{\partial \theta_i} (w, G(z_k, \theta), \gamma_k), \, \mbox{ for } \, i=1,\dots,q,\]
where
\[\frac{\partial\mathcal{L}}{\partial \theta_i} (w, G(z_k, \theta), \gamma_k) = w_i (\sigma(w, G(z_k, \theta)) - y_k) \frac{\partial G}{\partial \theta_i} (z_{ki}, \theta_i),\, \mbox{ for } \, i=1,\dots,q.\]
We define
\begin{equation}
\label{eqn:varsigma}
    \varsigma(w, \theta, z_k) := \sigma(w, G(z_k, \theta)) (1 - \sigma(w, G(z_k, \theta))), \, \mbox{ for } \, k = 1,\dots,m,
\end{equation}
and find the second derivative with respect to the parameters of the adversary's generator to be
\[\frac{\partial^2 F}{\theta_i \theta_j}(w, \theta) = \sum_{k=1}^n \frac{\partial\mathcal{L}}{\partial \theta_i \theta_j}(w, G(z_k, \theta), \gamma_k), \, \mbox{ for } \, i=1,\dots,q, \; j = 1,\dots,q,\]
where
\[\frac{\partial^2\mathcal{L}}{\partial \theta_i \theta_j} (w, G(z_k, \theta), \gamma_k) = \begin{cases} 
      w_i (\sigma(w, G(z_k, \theta)) - \gamma_k) 
\frac{\partial^2 G}{\partial \theta_i \theta_i} (z_{ki}, \theta_i) + w_i^2 \varsigma(w, \theta, z_k) \left ( \frac{\partial G}{\partial \theta_i} (z_{ki}, \theta_i) \right )^2 & \mbox{if }\, i=j, \\
      w_i w_j \varsigma(w, \theta, z_k) \frac{\partial G}{\partial \theta_i} (z_{ki}, \theta_i) \frac{\partial G}{\partial \theta_j} (z_{ki}, \theta_j) & \mbox{otherwise},
   \end{cases}\]
for $i=1,\dots,q,j = 1,\dots,q$, where $\varsigma$ is defined as in \eqref{eqn:varsigma}. Similarly, we find the joint derivative with respect to $w$ and $\theta$ element-wise to be
\[\frac{\partial^2 F}{w_l \theta_j} (w, \theta) = \sum_{k=1}^n \frac{\partial^2\mathcal{L}}{\partial w_l \theta_j} (w, G(z_k, \theta), \gamma_k), \, \mbox{ for } \, l = 1,\dots,q, \; j = 1,\dots,q,\]
where
\[\frac{\partial^2\mathcal{L}}{\partial w_l \theta_j} (w, G(z_k, \theta), \gamma_k) = \begin{cases} 
      \left ( (\varsigma(w, \theta, z_k) - y_k) + w_l G(z_{ki}, \theta_i) \varsigma(w, \theta, z_k) \right ) \frac{\partial G}{\partial \theta_i} (z_{ki}, \theta_i) & \mbox{ if }\, i=j, \\
      w_j \varsigma(w, \theta, z_k) G(z_{ki}, \theta_i) \frac{\partial G}{\partial \theta_j} (z_{ki}, \theta_j) & \mbox{ otherwise},
   \end{cases}\]
for $i=1,\dots,q,j = 1,\dots,q$, where $\varsigma$ is defined as in \eqref{eqn:varsigma}. 

The derivative of the lower-level objective with respect to the learner's weights is found to be
\[\nabla_w f(w, \theta) = G(z, \theta)^T(\sigma(w, G(z, \theta)) - (1 - \gamma)),\]
and the second derivative is found to be
\[\nabla^2_{ww} f(w, \theta) = \frac{1}{m} \tilde{G}(z, \theta) D_G(I_m - D_G) \tilde{G}(z, \theta)^T.\]
We define the derivatives with respect to the parameters of the adversary's generator element-wise, and find that
\[\frac{\partial f}{\partial \theta_i} (w, \theta) = \sum_{k=1}^n \frac{\partial \mathcal{L}}{\partial \theta_i} (w, G(z_k, \theta), 1 - \gamma_k), \, \mbox{ for } \, i=1,\dots,q.\]
The second derivatives of the lower-level objective with respect to the parameters of the adversary's generator are 
\[\frac{\partial^2 f}{\partial \theta_i \theta_j} (w, \theta) = \sum_{k=1}^n \frac{\partial^2 \mathcal{L}}{\partial \theta_i \theta_j} (w, G(z_k, \theta), 1 - \gamma_k), \, \mbox{ for } \, i=1,\dots,q, \; j=1,\dots,q.\]
Finally, the joint derivativse with respect to the learner's weights and the parameters of the adversary's generator are obtained as
\[\frac{\partial^2 f}{\partial w_l \theta_j} (w, \theta) = \sum_{k=1}^n \frac{\partial^2 \mathcal{L}}{\partial w_l \theta_j} (w, G(z_k, \theta), 1 - \gamma_k), \; \mbox{ for } \, w=1,\dots,q, \, i=1,\dots,q.\]
 \end{APPENDIX}

\end{document}